\documentclass[conference]{IEEEtran}
\ifCLASSINFOpdf
  \usepackage[pdftex]{graphicx}
\else
\fi
\usepackage[nocompress]{cite}

\usepackage{amsmath}
\usepackage{amsfonts}
\usepackage{mathtools,xparse}
\usepackage[section]{placeins}

\DeclarePairedDelimiter{\norm}{\lVert}{\rVert}
\NewDocumentCommand{\normL}{ s O{} m }{%
  \IfBooleanTF{#1}{\norm*{#3}}{\norm[#2]{#3}}_{L_2(\Omega)}%
}

\begin{document}
%
\title{Towards Radiologist-Level Accurate Deep Learning
System for Pulmonary Screening}

\author{\IEEEauthorblockN{Mrinal Haloi, Raja Rajalakshmi K, Pradeep Walia}
\IEEEauthorblockA{Artelus\\
\{mrinalhaloi, rkodhandapani, pwalia\}@artelus.com}
}


%


\maketitle

\begin{abstract}
In this work, we propose advanced pneumonia and Tuberculosis grading system for X-ray images. The proposed system is a very deep fully convolutional classification network with online augmentation that outputs confidence values
for diseases prevalence. It’s a fully automated system capable of disease feature understanding without any offline preprocessing step or manual feature extraction. We have achieved state-
of-the-art performance on the public databases such as ChestXray-14, Mendeley,  Shenzhen Hospital X-ray and Belarus X-ray set.
\end{abstract}


%
\IEEEpeerreviewmaketitle

\section{Introduction}
One-third of the world's population is thought to be infected with TB \cite{tb_1}. New infections occur in about 1\% of the population each year \cite{tb_2}. In 2016, there were more than 10 million cases \cite{tb_3} of active TB which resulted in 1.3 million deaths. This makes it the number one cause of death from an infectious disease and more than 95\% of deaths occurred in developing countries. Pneumonia affects approximately 450 million people \cite{np_1} globally (7\% of the population) and results in about 4 million deaths per year.

\begin{figure}[h]
  \centering
      \includegraphics[width=3.0in,height=1.7in]{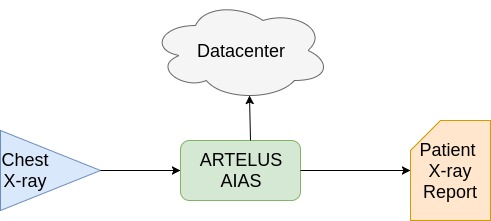}
\caption{AIAS system for X-ray screening}
\label{fig:aias}
\end{figure}

For screening and diagnosis of Pneumonia and Tuberculosis chest X-rays \cite{xray} play a very critical role due to its availability and affordability. Expert radiologists are required to interpret results from X-rays and it's a challenging task. There aren’t an adequate number of
experienced radiologists especially in the developing countries, many patients don’t get proper care. Fast and reliable automatic computer-aided screening
and diagnosis system will reduce the burden on specialists and will give better performance for mass screening. An automated artificial intelligence aided screening (AIAS) system shown in Fig.~\ref{fig:aias} will help patients at remote areas where radiologists and connectivity aren’t readily accessible. 

\begin{figure}[h]
  \centering
      \includegraphics[width=3.0in,height=2.7in]{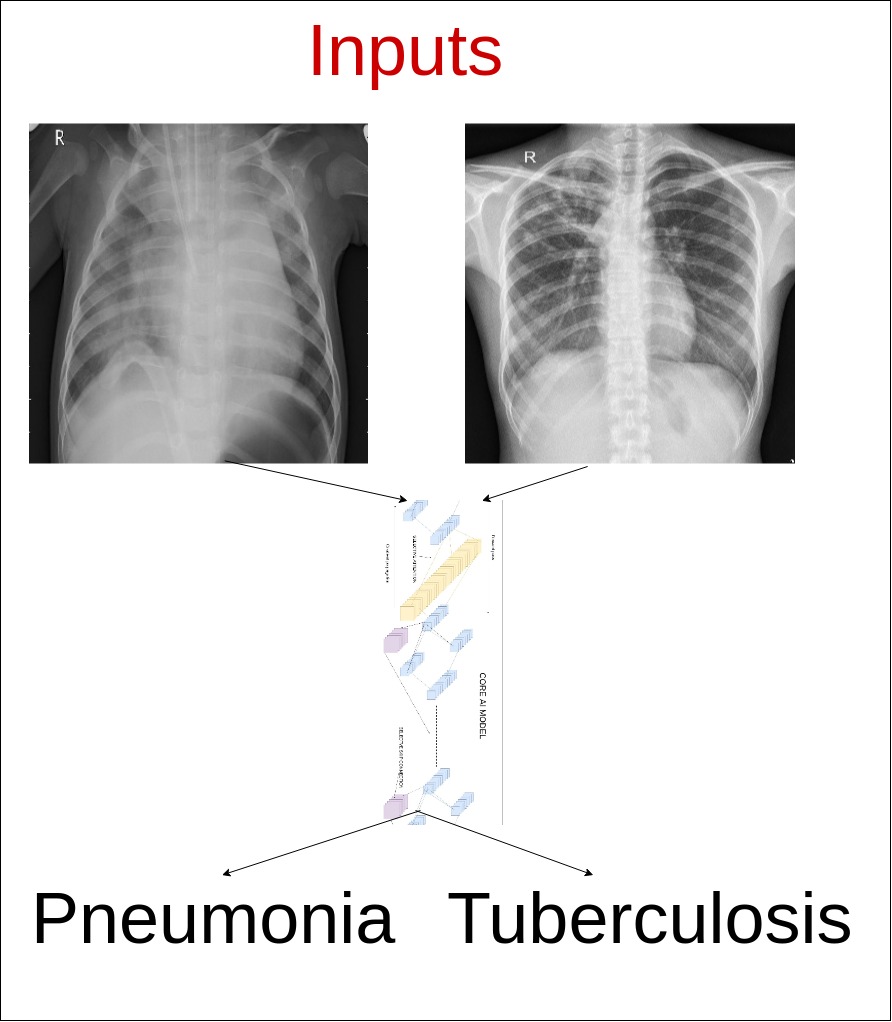}
\caption{Chest X-ray: Pneumonia and Tuberculosis cases; using our 211 layers deep model the above cases are correctly identified}
\label{fig:intro}
\end{figure}

Classification of Pneumonia and Tuberculosis is a challenging task due to many similar pathological features associated with other diagnoses. Even expert radiologists make wrong diagnoses due to complex nature of these features. This results in disagreements even among radiologists\cite{npconflict} for Pneumonia and TB classification. 

We make use of deep learning to develop an advanced X-ray screening system. Deep learning is an algorithmic technique
that is revolutionizing what is possible in areas such as finance,
healthcare, communication, automotive, natural language processing, computer vision and more. It allows computers to
analyze vast amounts of data and automatically detect patterns
and make accurate predictions. Deep Learning could help
doctors screening and diagnosis Pneumonia and Tuberculosis more quickly and more accurately.

In most of the computer-based screening environment setting sensitivity and specificity are used as AIAS system's efficiency measurement criteria. Using our AIAS system we report 96 \% sensitivity for Pneumonia and 92.5 \% sensitivity for Tuberculosis on publicly available datasets.

\subsection{Contribution}
\textbf{X-ray classification learning:} We present a radiologists performance level chest X-ray AIAS system for Tuberculosis and Pneumonia.
  Following are the advantages of the proposed AIAS system:
    \begin{itemize}
    \item Trained on a small number of samples with better generalization performance.
    \item State-of-the-art results for Tuberculosis and Pneumonia classification.
    \item Faster and deployable in a mobile chip.
    \item Accessible healthcare in the remote areas. 
    \end{itemize}

\section{Method}

\subsection{Dataset}
We have used four publicly available datasets in this work.
\begin{itemize}
\item ChestXray-14 \cite{chestxray14}: The NIH Clinical Center published this dataset of high-quality X-ray images of 32K unique patients. It includes 112K images and 14 associated diseases labels mined from radiologists report using natural language processing. 

\item Mendeley \cite{mendeley}: Mendeley published this dataset of 5856 X-ray images of patients diagnosed with Pneumonia. This set contains both normal cases and cases with manifestations of Pneumonia. 

\item MC-SC-Xray \cite{chinaset}: It includes datasets from Montgomery County chest X-ray set and Shenzhen chest X-ray set. The set contains 800 X-ray images, of which 406 are normal cases and 394 are cases with manifestations of TB.

\item Belarus \cite{belarus}: Belarus public health published X-ray images of 107 patients images. The set contains both normal cases and cases with manifestations of TB.
 
\end{itemize}

\subsection{Proposed Algorithm} Deep learning is proved to be the current state-of-the-art for
computer vision/image processing, speech, text problems. We
propose a very deep network to predict disease label based on
X-ray image pixel values. The proposed network has around 45
million trainable parameters and 211 layers deep. The proposed
network is a modified version of network architecture \cite{haloi}. Values of those parameters are learned using stochastic gradient descent with Nesterov momentum
algorithm. All parameters are initialized using random values
sampled from a truncated Gaussian distribution. The parameter of learning is an iterative process based on disagreements between
the grades predicted by the network and the radiologists.
The disagreements value is used as feedback to optimize
network parameters to understand chest pathological features.

\subsection{Network Architecture}

\begin{figure}[h]
  \centering
      \includegraphics[width=3.1in,height=2.1in]{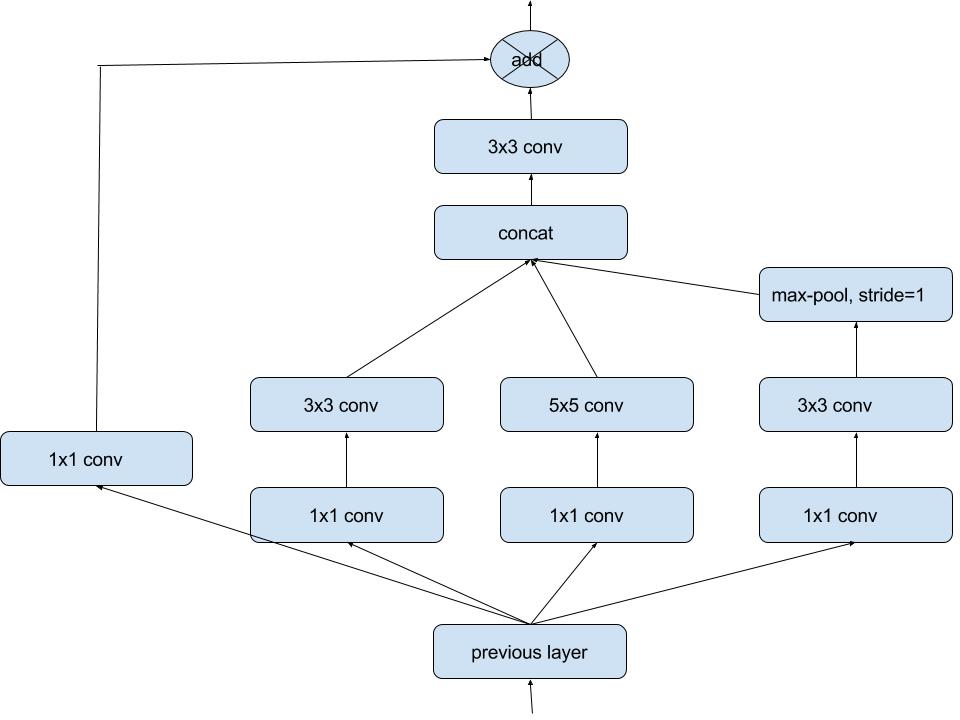}
\caption{Residual Inception}
\label{fig:incept}
\end{figure}

\begin{figure}[h]
  \centering
      \includegraphics[width=3.1in,height=2.1in]{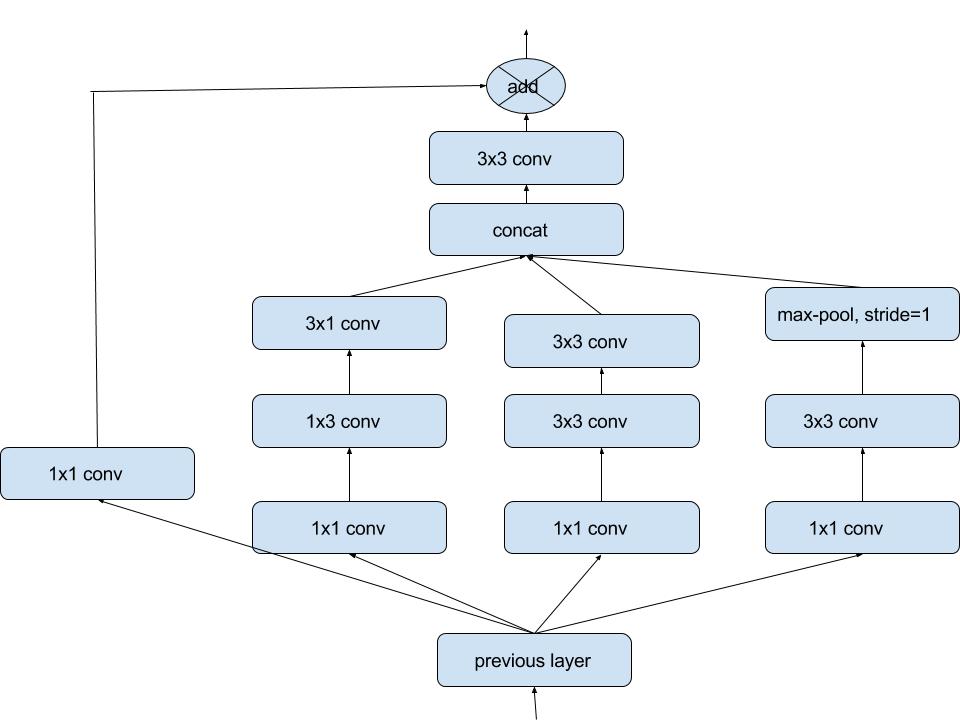}
\caption{Residual Inception with convolution factorization for the classifier node}
\label{fig:incept2}
\end{figure}

\textbf{Residual Inception module} We have used a modified version of the base inception block proposed in \cite{haloi}, where block-based convolution layers and skip connections were used to ensure rich feature representation. We have added a residual connection from the block input to output as shown in Fig~\ref{fig:incept}. The concatenated features were convolved with another $3\times3$ convolution to reduce aliasing and added with $1\times1$ convolved input features; both operations have the same number of output filters. Apart from that, we have used a convolution factorization version Fig~\ref{fig:incept2} of the same module for the classifier.


\begin{figure}[h]
  \centering
      \includegraphics[width=3.1in,height=1.7in]{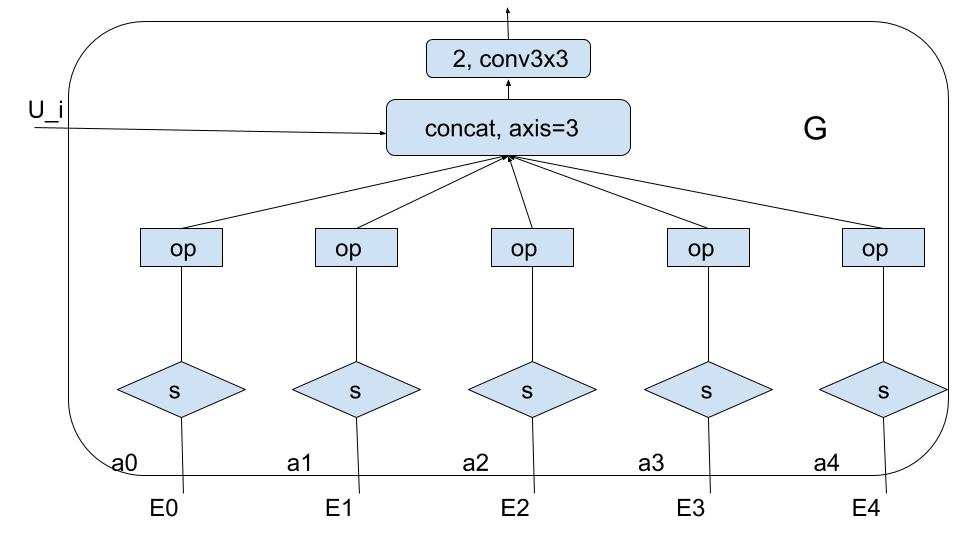}
\caption{Partial attention over encoder}
\label{fig:attention}
\end{figure}

\textbf{Partial attention}
We use partial attention \cite{rethink} over the selected classifier feature maps. Feeding the intermediate feature maps information to boost the reconstructed feature map representation power. The attention mechanism here only focuses on the feature maps with spatial resolutions equal to or larger than that of the target feature map outputs. Max-pooling to reduce spatial resolution of encoder feature maps to the target output size; max-pooling ensures maximally activated features. Fig~\ref{fig:attention} shows an overview of the partial attention mechanism used, where $E_i$ are the last feature maps (before reducing the feature map width and height for each block) of each convolution block of the encoder. The switch $s$ is on if the corresponding $a_i > 0$; $op$ is a max-pooling operation for all $a_i >1$ with stride $a_{i}$ and identity operation for $a_i=1$. 
$U_i$ is the feature representation of the target layer.
\begin{equation}
a_i = \frac{min(height_{E_{i}}, width_{E_{i}})}{min(height_{U_{i}}, width_{U_{i}})}
\end{equation}

\section{EXPERIMENTS AND RESULTS}
\textbf{Dataset:} We have used ChestXray-14 dataset (80\% of the total dataset cases) to pre-train the network. The training dataset has around 89K cases with 14 distinct labels. We have gathered the finetuning set of Tuberculosis and Pneumonia cases using ChestXray-14 held-out set (20\% of the total dataset cases), Mendeley, MC-SC-Xray and Belarus datasets; it includes 20K training cases and 1.8K validation cases.

\textbf{Metrics:} The network performance was measured using AUC, sensitivity and specificity. AUC is the area under the receiver operating characteristics curve (ROC). 
Sensitivity or true positive rate (TPR) is the percentage of the pathological samples that are classified correctly, defined in eq~\ref{eq:tpr}. Specificity of true negative rate is the percentage of the normal samples that are classified correctly, defined in eq~\ref{eq:tnr}.
\begin{equation}
\label{eq:tpr}
TPR = sensitivity = \frac{TP}{TP + FN}
\end{equation}
\begin{equation}
\label{eq:tnr}
TPR = sensitivity = \frac{TN}{FP + TN}
\end{equation}

\textbf{Transfer Learning:}
We have used transfer learning on the finetuning set. The pre-trained model on ChestXray-14 dataset learns diverse and discriminative features of chest X-ray images. Using this pre-trained model we fine-tune the model for Tuberculosis and Pneumonia classification.

\textbf{Preprocessing:}
For the classifier, $224\times224\times3$ is used as the input size. All input images were resized to $256\times256\times3$ and a randomly cropped patch of size $224\times224\times3$ is used as input. Extensive data augmentation is used such as random flip left/right, up/down and changing the image pixel values randomly using hue, contrast, and saturation. Also, each image is standardized by it's mean and dividing its standard deviation.   

\textbf{Framework:} We have used TEFLA\cite{tefla}, a python framework developed on the top of TENSORFLOW\cite{tf}, for all the experiments described in this work.

\begin{figure}[h]
  \centering
      \includegraphics[width=3.1in,height=2.1in]{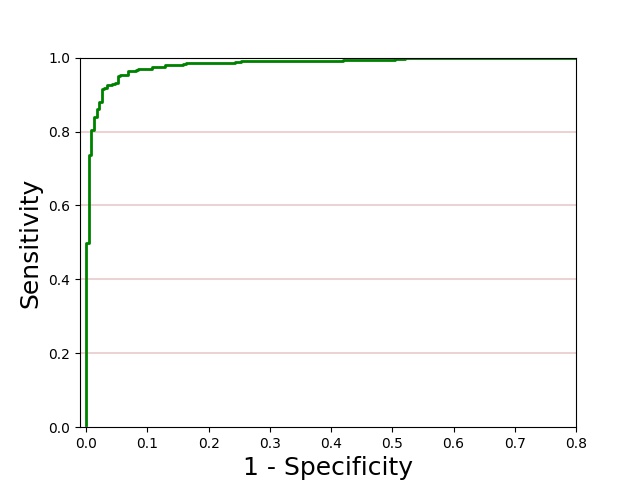}
\caption{ROC plot for Pneumonia with 0.985 AUC; by changing the prediction threshold we can tune the sensitivity and the specificity for various requirements.}
\label{fig:np}
\end{figure}

\begin{figure}[h]
  \centering
      \includegraphics[width=3.1in,height=2.1in]{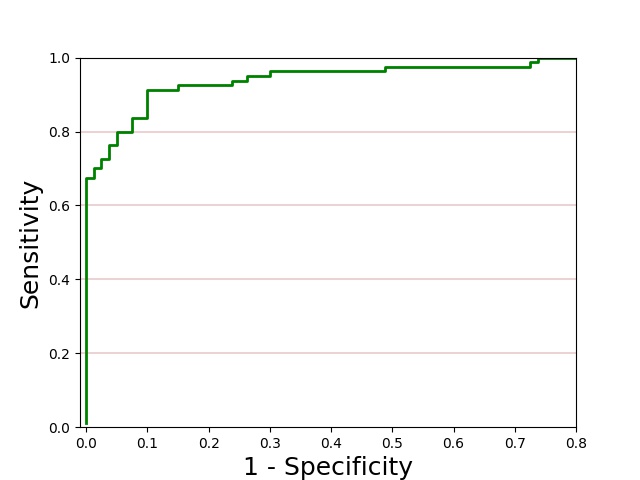}
\caption{ROC plot for Tuberculosis with 0.949 AUC; by changing the prediction threshold we can tune the sensitivity and the specificity for various requirements.}
\label{fig:tb}
\end{figure}
\begin{table}[h]
\begin{center}
  \begin{tabular}{ | p{3cm} | c |c|c| }
    \hline
    Diseases & Sensitivity (\%) & Specificity (\%) & AUC \\ \hline
    Pneumonia & 96.1 & 91.03 & 0.985 \\ 
    \hline
    Tuberculosis & 92.5 & 85.0 & 0.949 \\
    \hline
  \end{tabular}
\end{center}
\caption{Model performance for Pneumonia and TB classification}
\label{table:comp}
\end{table}
\textbf{Training procudure:} We formulate the AIAS system as multi-label classification problem in which each class is independent and not mutually exclusive. For our scenario, a chest X-ray can contain the pathological features of both the TB and Pneumonia at the same time. Weighted sigmoid cross entropy loss best suited for problems with overlapping domains is used to optimize the model parameters. Eq~\ref{eq:loss} outlines the weighted cross entropy loss in which $n$ denotes the number of classes and $w_{c}$ denotes the weight of the class $c$. For generalization various regularization approaches \cite{general} can be used. Batch normalization\cite{bn} is used to reduce covariate shift and achieve faster convergence of both models. Also, we have used Nesterov momentum optimizer with polynomial learning rate policy. Gradient normalization\cite{gn} was used to stabilize the training process. To improve generalization capability we
have used label smoothing (soft targets). Dropout and batch
sample balancing technique was used to ensure that network doesn't overfit. 
Five models with different
architectures were trained on the same training and
validation set. For each model, we have also explored
weight initializations from the same network trained on
the Imagenet \cite{imagenet} dataset.

\begin{equation}
\label{eq:loss}
\begin{aligned}
Loss = - \sum_{c=0}^{c=n} w_{c} y_{true}log(y_{predicted}) \\
\end{aligned}
\end{equation}

\begin{figure}[h]
  \centering
      \includegraphics[width=3.0in,height=2.8in]{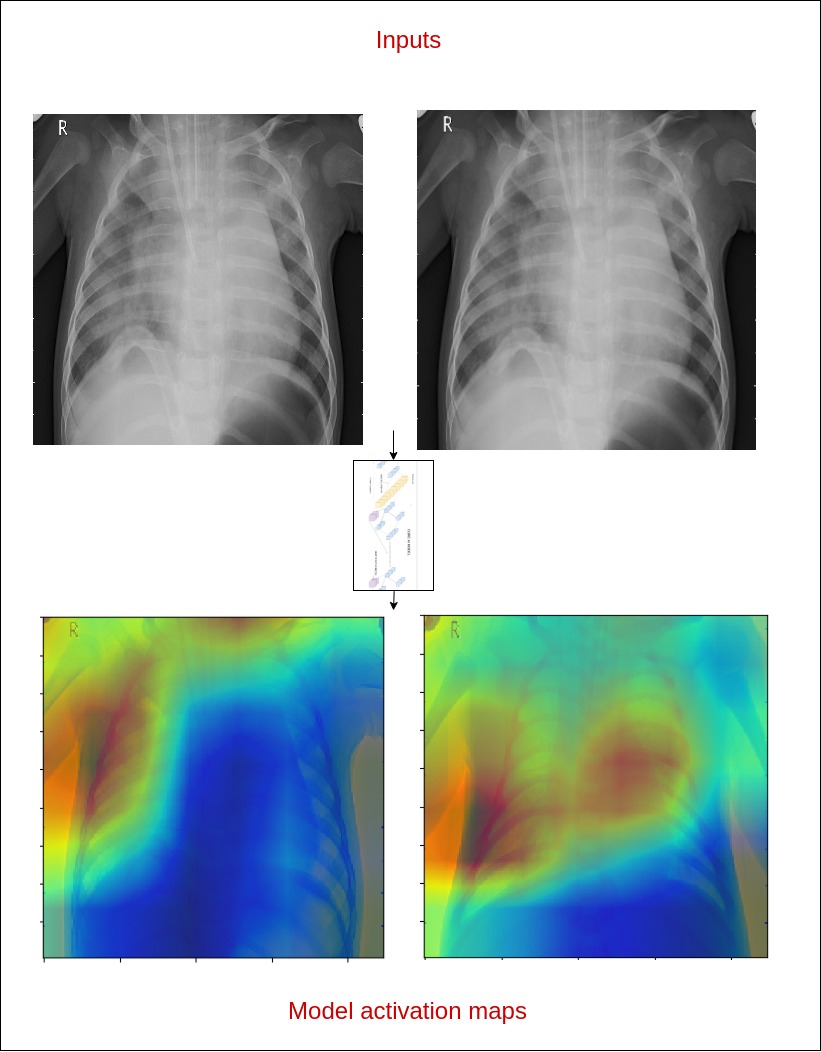}
\caption{Heatmap showing the model's areas of interest while predicting diseases. Here the model focuses at the correct locations with pathological features; this can be used for localization.}
\label{fig:heatmap}
\end{figure}
\begin{table}[h]
\begin{center}
  \begin{tabular}{ | p{3cm} | c |c|c| }
    \hline
    Method & Sensitivity (\%) & Specificity (\%) & AUC \\ \hline
    \textbf{Proposed Model} & 92.5 & 85.0 & 0.949 \\
    \hline
    Jaeger\cite{jeager} & \_ & \_ & 0.900 \\
    \hline
    Hwang\cite{hwang} & \_ & \_ & 0.93 \\
    \hline
    MTIslam\cite{mtislam} & 80.0 & 92.0 & 0.91 \\
    \hline
  \end{tabular}
\end{center}
\caption{Model performance comparisons for Tuberculosis}
\label{table:comp_pn}
\end{table}

\begin{table}[h]
\begin{center}
  \begin{tabular}{ | p{3cm} | c |c|c| }
    \hline
    Method & Sensitivity (\%) & Specificity (\%) & AUC \\ \hline
    \textbf{Proposed Model} & 96.1 & 91.03 & 0.985 \\ \hline
    Wang\cite{chestxray14} & \_ & \_ & 0.633 \\ \hline
    Yao\cite{yao} & \_ & \_ & 0.713 \\ \hline
    CheXNet\cite{chexnet} & \_ & \_ & 0.768 \\ \hline
  \end{tabular}
\end{center}
\caption{Model performance comparisons for Pneumonia}
\label{table:comp_tb}
\end{table}

Figure~\ref{fig:np} shows the receiver operating curve (AUC) for Pneumonia classification and Figure~\ref{fig:tb} shows the receiver operating curve for Tuberculosis classification. We have surpassed radiologists level performance. 
Table~\ref{table:comp} shows the model performance on the dataset with Tuberculosis and Pneumonia cases. We have achieved 96\% sensitivity and 91\% specificity for Pneumonia classification with 0.93 AUC. Tuberculosis we have achieved 92.5\% sensitivity and 85\% specificity with 0.89 AUC.  

Extensive performance comparisons among several Pneumonia classification methods and the proposed deep model is shown in Table~\ref{table:comp_pn}. Our model performs significantly better than the current state-of-the-art approaches. Table~\ref{table:comp_tb} shows the Tuberculosis classification performance comparisons of our model and several other methods.

The feature representations learning capacity of the X-ray classification deep model can be understood by its ability to pay attention at the input image regions with pathological features.  Fig.~\ref{fig:heatmap} shows the image regions the model looks at while predicting disease label.

\subsection{Limitations}
The AIAS system is trained on the frontal chest X-ray images; which results in the system inability to accurately diagnosis radiographs with lateral views. Overexposed or underexposed chest radiographs affect the radiologist's readings and also that of our system. Patients medical histories are not taken into account while designing the model. 

\section{Conclusion}
In this work of chest X-rays images, a deep learning
based AIAS system with high sensitivity and specificity for
detecting Pneumonia and Tuberculosis is proposed. The proposed AIAS system with sensitivity 96\% will enhance  (accuracy and results time) for Pulmonary screening performance. Temporal consistency of grading across X-ray images for a
specific operating point is an added efficiency of an automated
AIAS system. We aim to incorporate radiographs with lateral views in our next version of the model also, we will integrate patient medical histories and lifestyle information to have added advantages to the model while predicting.



%

\section{Appendix}

\subsection{Results on ChestXray-14 Test set}	
We have also computed the test performance of the model pre-trained on the ChestX-ray-14 dataset. Table~\ref{table:compchextray} shows the performance comparisons for different abnormalities among various existed methods and the proposed model. We outperform recent state-of-the-art methods on the ChestXray-14 test set.  Deep learning can improve and help radiologists in the diagnosis and screening of diseases associated with the chest. It will also reduce response time to generate reports.
\begin{table}[h]
\begin{center}
  \begin{tabular}{ | p{1.5cm} | c |c|c| p{1.2cm}|}
    \hline
    Abnormalities & Wang\cite{chestxray14} & Yao\cite{yao} & Chexnet\cite{chexnet} & \textbf{Proposed Method} \\ \hline
    
 Atelectasis & 0.716 & 0.772 & 0.8094 & 0.857 \\ \hline
Cardiomegaly & 0.807 & 0.904 & 0.9248 & 0.918 \\ \hline
Effusion & 0.784 & 0.859 & 0.8638 & 0.903 \\ \hline 
Infiltration & 0.609 & 0.695 & 0.7345 & 0.912 \\ \hline
Mass & 0.706 & 0.792 & 0.8676 & 0.873 \\ \hline
Nodule & 0.671 & 0.717 & 0.7802 & 0.840 \\ \hline
Pneumonia & 0.633 & 0.713 & 0.7680 & 0.912 \\ \hline
Pneumothorax & 0.806 & 0.841 & 0.8887 & 0.906 \\ \hline
Consolidation & 0.708 & 0.788 & 0.7901 & 0.871 \\ \hline 
Edema & 0.835 & 0.882 & 0.8878 & 0.911 \\ \hline
Emphysema & 0.815 & 0.829 & 0.9371 & 0.925 \\ \hline
Fibrosis & 0.769 & 0.767 & 0.8047 & 0.891 \\ \hline
Pleural Thickening & 0.708 & 0.765 & 0.8062 & 0.901 \\ \hline
Hernia & 0.767 & 0.914 & 0.9164 & 0.921 \\ \hline   
 
  \end{tabular}
\end{center}
\caption{The pre-trained model AUC comparisons with existed methods}
\label{table:compchextray}
\end{table}

\subsection{MODS Assay Test For TB Validation}
The ability to do faster more accurate diagnosis leads us to our primary goal of reducing the time it takes to screen population in the vulnerable areas or poor socio-economic zones for early diagnosis of TB and to minimize the spreading of the germs and transmitting and infecting others in the group. The ultimate goal of our research is to reduce patient wait times for being diagnosed with this infectious disease by developing new machine learning and mobile health techniques to the TB screening and diagnosis problem.

This approach is greatly in contrasts with the global health community which has focused its efforts on developing and testing effective vaccines, improving the diagnosis process, and promoting patient compliance to the medical treatment.

Efforts to eliminate the TB epidemic are challenged by the persistent social inequalities in health, the small number of local healthcare professionals, and the weak healthcare infrastructure found in resource-poor settings.

Specifically, we aim to design a user-centered, mobile device-based computing system to significantly expedite the TB diagnosis process by Developing novel image processing and machine learning Techniques to analyze chest X-ray images.

Once a candidate is suspected of having TB during the xray examination, a MODS test will be administered,  and if positive, appropriate medication will be started, and appropriate treatment pathway and protocol will be followed.

\subsection{MODS Assay Test}

The “MODS’(Microscopic -Observation Drug Susceptibility) assay is a faster, low cost, high performance, rapid alternative for conventional methods for drug susceptibility testing of Mycobacterium tuberculosis and DST (drug susceptibility testing) directly from a sputum sample which was developed by a research team in Lima, Peru.

The MODS assay is based on three principles:
\begin{itemize}

\item Mycobacterium tuberculosis (MTB) grows faster in liquid media than on solid media.

\item  Microscopic MTB growth can be detected earlier in liquid media than waiting for the macroscopic appearance of colonies on solid media, and that growth is characteristic of MTB, allowing it to be distinguished from atypical mycobacteria or fungal or bacterial contamination.

\item The drugs isoniazid and rifampicin can be incorporated into the MODS assay to allow for simultaneous direct detection of MDRTB (multidrug resistant tuberculosis), obviating the need for subculture to perform an indirect drug susceptibility test.
\end{itemize}

The underlying philosophy of MODS is laboratory freeware, all the components are readily available from laboratory suppliers.

MODS has also been recommended by the World Health Organization (WHO) as an affordable and highly effective alternative to existing gold standard liquid mycobacterial culture methods for testing sputum samples of TB-suspected individuals. This is a very useful screening tool to combat the great burden of the disease in developing countries.

\end{document}